\documentclass{article}

\PassOptionsToPackage{numbers, compress}{natbib}
\usepackage[final]{nips_2016}

\usepackage[utf8]{inputenc} 
\usepackage[T1]{fontenc}    
\usepackage{hyperref}       
\usepackage{url}            
\usepackage{booktabs}       
\usepackage{amsfonts}       
\usepackage{nicefrac}       
\usepackage{microtype}      

\usepackage{tkz-graph}
\usetikzlibrary{positioning}

\usepackage{times}
\usepackage{graphicx} 
\usepackage{caption}
\usepackage{subcaption}

\usepackage{amsmath}
\usepackage{amsfonts}
\usepackage{amssymb}
\usepackage{amsthm}

\renewcommand{\c}{\mathbf{c}}

\usepackage[utf8]{inputenc}
\usepackage{amsthm}
\usepackage{amsmath}
\usepackage{amsfonts}
\usepackage{amssymb}
\usepackage{dsfont}
\usepackage{color}

\allowdisplaybreaks

\newcommand{\N}{\mathcal{N}}
\newcommand{\cL}{\mathcal{L}}

\newcommand{\td}{\text{d}}
\newcommand{\f}{\mathbf{f}}
\newcommand{\x}{\mathbf{x}}
\newcommand{\Bb}{\mathbf{b}}

\newcommand{\y}{\mathbf{y}}

\newcommand{\w}{\mathbf{w}}
\newcommand{\W}{\mathbf{W}}

\newcommand{\m}{\mathbf{m}}

\newcommand{\X}{\mathbf{X}}
\newcommand{\Y}{\mathbf{Y}}

\newcommand{\I}{\mathbf{I}}
\newcommand{\M}{\mathbf{M}}

\newcommand{\bz}{\mathbf{0}}

\newcommand{\bo}{\text{\boldmath$\omega$}}

\newcommand{\KL}{\text{KL}}

\theoremstyle{definition}

\def\fl[#1\]{\begin{align}#1\end{align}}
\def\[#1\]{\begin{align*}#1\end{align*}}

\def\*[#1\]{\begin{align*}#1\end{align*}}

\usepackage{dblfloatfix}

\title{A Theoretically Grounded Application of Dropout in Recurrent Neural Networks}

\author{
\And
Yarin Gal 
\And
\\
University of Cambridge\\
\texttt{\{yg279,zg201\}@cam.ac.uk} 
\And
Zoubin Ghahramani
}

\begin{document} 

\maketitle

\begin{abstract}
Recurrent neural networks (RNNs) stand at the forefront of many recent developments in deep learning. Yet a major difficulty with these models is their tendency to overfit, with dropout shown to fail when applied to recurrent layers. 
Recent results at the intersection of Bayesian modelling and deep learning offer a Bayesian interpretation of common deep learning techniques such as dropout.
This grounding of dropout in approximate Bayesian inference suggests an extension of the theoretical results, offering insights into the use of dropout with RNN models.
We apply this new variational inference based dropout technique in LSTM and GRU models, assessing it on language modelling and sentiment analysis tasks. The new approach outperforms existing techniques, and to the best of our knowledge improves on the single model state-of-the-art in language modelling with the Penn Treebank (73.4 test perplexity).
This extends our arsenal of variational tools in deep learning.
\end{abstract} 

\section{Introduction}

Recurrent neural networks (RNNs) are sequence-based models of key importance for natural language understanding, language generation, video processing, and many other tasks \citep{sundermeyer2012lstm,
kalchbrenner2013recurrent,
sutskever2014sequence}. 
The model's input is a sequence of symbols, where at each time step a simple neural network (\textit{RNN unit}) is applied to a single symbol, as well as to the network's output from the previous time step.
RNNs are powerful models, showing superb performance on many tasks, but overfit quickly. 
Lack of regularisation in RNN models makes it difficult to handle small data, and to avoid overfitting researchers often use early stopping, or small and under-specified models \citep{zaremba2014recurrent}.

Dropout is a popular regularisation technique with  deep networks \citep{hinton2012improving,
srivastava2014dropout} where network units are randomly masked during training (\textit{dropped}).
But the technique has never been applied successfully to RNNs. 
Empirical results have led many to believe that noise added to recurrent layers (connections between RNN units) will be amplified for long sequences, and drown the signal \citep{zaremba2014recurrent}. 
Consequently, existing research has concluded that the technique should be used with the inputs and outputs of the RNN alone \citep{zaremba2014recurrent,
pachitariu2013regularization,
bayer2013fast,
pham2014dropout,
bluche2015apply}. But this approach still leads to overfitting, as is shown in our experiments. 

Recent results at the intersection of Bayesian research and deep learning offer interpretation of common deep learning techniques through Bayesian eyes \citep{rezende2014stochastic,
blundell2015weight,
hernandez2015probabilistic,
Gal2015Bayesian,
Kingma2015,
Murphy2015}. This Bayesian view of deep learning allowed the introduction of new techniques into the field, such as methods to obtain principled uncertainty estimates from deep learning networks \citep{Gal2015Bayesian,Gal2015DropoutB}.
\citet{Gal2015Bayesian} for example showed that dropout can be interpreted as a variational approximation to the posterior of a Bayesian neural network (NN). Their variational approximating distribution is a mixture of two Gaussians with small variances, with the mean of one Gaussian fixed at zero.
This grounding of dropout in approximate Bayesian inference suggests that an extension of the theoretical results might offer insights into the use of the technique with RNN models.

Here we focus on common RNN models in the field (LSTM \citep{hochreiter1997long}, GRU \citep{cho2014Learning}) and interpret these as probabilistic models, i.e.\ as RNNs with network weights treated as random variables, and with suitably defined likelihood functions.
We then perform approximate variational inference in these probabilistic Bayesian models (which we will refer to as \textit{Variational RNNs}). Approximating the posterior distribution over the weights with a mixture of Gaussians (with one component fixed at zero and small variances) will lead to a tractable optimisation objective. 
Optimising this objective is identical to performing a new variant of dropout in the respective RNNs.

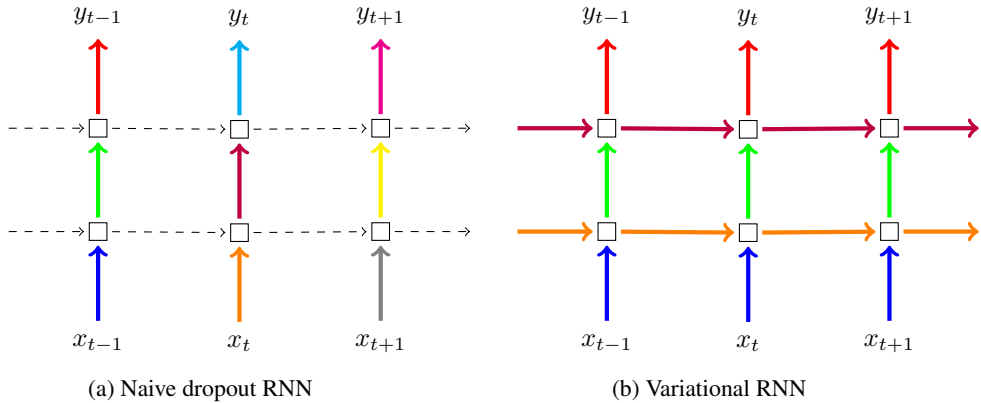
\begin{figure*}[t!]
\center
\hspace{-12mm}
\begin{subfigure}[t]{0.4\linewidth}
\center
\begin{tikzpicture}
\node[align=center, outer sep=2] (x_t) {$x_t$};
\node[draw, align=center, outer sep=2] (unit_t1) [above=of x_t] {};
\node[draw, align=center, outer sep=2] (unit_t2) [above=of unit_t1] {};
\node[align=center, outer sep=2] (y_t) [above=of unit_t2] {$y_t$};

\node[align=center, outer sep=2] (x_tm1) [left=of x_t] {$x_{t-1}$};
\node[draw, align=center, outer sep=2] (unit_tprev1) [above=of x_tm1] {};
\node[draw, align=center, outer sep=2] (unit_tprev2) [above=of unit_tprev1] {};
\node[align=center, outer sep=2] (y_tm1) [above=of unit_tprev2] {$y_{t-1}$};

\node[align=center, outer sep=2] (x_tp1) [right=of x_t] {$x_{t+1}$};
\node[draw, align=center, outer sep=2] (unit_tnext1) [above=of x_tp1] {};
\node[draw, align=center, outer sep=2] (unit_tnext2) [above=of unit_tnext1] {};
\node[align=center, outer sep=2] (y_tp1) [above=of unit_tnext2] {$y_{t+1}$};

\node (unit_tnextnext1) [right=of unit_tnext1] {};
\node (unit_tnextnext2) [right=of unit_tnext2] {};
\node (unit_tprevprev1) [left=of unit_tprev1] {};
\node (unit_tprevprev2) [left=of unit_tprev2] {};

\path[->, draw=blue, ultra thick] (x_tm1) edge (unit_tprev1);
\path[->, draw=orange, ultra thick] (x_t) edge (unit_t1);
\path[->, draw=gray, ultra thick] (x_tp1) edge (unit_tnext1);
\path[->, draw=green, ultra thick] (unit_tprev1) edge (unit_tprev2);
\path[->, draw=purple, ultra thick] (unit_t1) edge (unit_t2);
\path[->, draw=yellow, ultra thick] (unit_tnext1) edge (unit_tnext2);
\path[->, draw=red, ultra thick] (unit_tprev2) edge (y_tm1);
\path[->, draw=cyan, ultra thick] (unit_t2) edge (y_t);
\path[->, draw=magenta, ultra thick] (unit_tnext2) edge (y_tp1);
\path[->, dashed] (unit_tprevprev1) edge (unit_tprev1);
\path[->, dashed] (unit_tprev1) edge (unit_t1);
\path[->, dashed] (unit_t1) edge (unit_tnext1);
\path[->, dashed] (unit_tnext1) edge (unit_tnextnext1);
\path[->, dashed] (unit_tprevprev2) edge (unit_tprev2);
\path[->, dashed] (unit_tprev2) edge (unit_t2);
\path[->, dashed] (unit_t2) edge (unit_tnext2);
\path[->, dashed] (unit_tnext2) edge (unit_tnextnext2);
\end{tikzpicture}
\caption{Naive dropout RNN}
\end{subfigure}
\hspace{10mm}
\begin{subfigure}[t]{0.4\linewidth}
\center
\begin{tikzpicture}
\node[align=center, outer sep=2] (x_t) {$x_t$};
\node[draw, align=center, outer sep=2] (unit_t1) [above=of x_t] {};
\node[draw, align=center, outer sep=2] (unit_t2) [above=of unit_t1] {};
\node[align=center, outer sep=2] (y_t) [above=of unit_t2] {$y_t$};

\node[align=center, outer sep=2] (x_tm1) [left=of x_t] {$x_{t-1}$};
\node[draw, align=center, outer sep=2] (unit_tprev1) [above=of x_tm1] {};
\node[draw, align=center, outer sep=2] (unit_tprev2) [above=of unit_tprev1] {};
\node[align=center, outer sep=2] (y_tm1) [above=of unit_tprev2] {$y_{t-1}$};

\node[align=center, outer sep=2] (x_tp1) [right=of x_t] {$x_{t+1}$};
\node[draw, align=center, outer sep=2] (unit_tnext1) [above=of x_tp1] {};
\node[draw, align=center, outer sep=2] (unit_tnext2) [above=of unit_tnext1] {};
\node[align=center, outer sep=2] (y_tp1) [above=of unit_tnext2] {$y_{t+1}$};

\node (unit_tnextnext1) [right=of unit_tnext1] {};
\node (unit_tnextnext2) [right=of unit_tnext2] {};
\node (unit_tprevprev1) [left=of unit_tprev1] {};
\node (unit_tprevprev2) [left=of unit_tprev2] {};

\path[->, draw=blue, ultra thick] (x_tm1) edge (unit_tprev1);
\path[->, draw=blue, ultra thick] (x_t) edge (unit_t1);
\path[->, draw=blue, ultra thick] (x_tp1) edge (unit_tnext1);
\path[->, draw=green, ultra thick] (unit_tprev1) edge (unit_tprev2);
\path[->, draw=green, ultra thick] (unit_t1) edge (unit_t2);
\path[->, draw=green, ultra thick] (unit_tnext1) edge (unit_tnext2);
\path[->, draw=red, ultra thick] (unit_tprev2) edge (y_tm1);
\path[->, draw=red, ultra thick] (unit_t2) edge (y_t);
\path[->, draw=red, ultra thick] (unit_tnext2) edge (y_tp1);
\path[->, draw=orange, ultra thick] (unit_tprevprev1) edge (unit_tprev1);
\path[->, draw=orange, ultra thick] (unit_tprev1) edge (unit_t1);
\path[->, draw=orange, ultra thick] (unit_t1) edge (unit_tnext1);
\path[->, draw=orange, ultra thick] (unit_tnext1) edge (unit_tnextnext1);
\path[->, draw=purple, ultra thick] (unit_tprevprev2) edge (unit_tprev2);
\path[->, draw=purple, ultra thick] (unit_tprev2) edge (unit_t2);
\path[->, draw=purple, ultra thick] (unit_t2) edge (unit_tnext2);
\path[->, draw=purple, ultra thick] (unit_tnext2) edge (unit_tnextnext2);
\end{tikzpicture}
\caption{Variational RNN}
\end{subfigure}

\vspace{4mm}
\caption{\textbf{Depiction of the dropout technique following our Bayesian interpretation (right) compared to the standard technique in the field (left).} Each square represents an RNN unit, with horizontal arrows representing time dependence (recurrent connections). Vertical arrows represent the input and output to each RNN unit. Coloured connections represent dropped-out inputs, with different colours corresponding to different dropout masks. Dashed lines correspond to standard connections with no dropout. Current techniques (naive dropout, left) use different masks at different time steps, with no dropout on the recurrent layers. The proposed technique (Variational RNN, right) uses the same dropout mask at each time step, including the recurrent layers.}
\label{fig:depiction}
\end{figure*}

In the new dropout variant, we repeat the same dropout mask at each time step for both inputs, outputs, and recurrent layers (drop the same network units at each time step). This is in contrast to the existing ad hoc techniques where different dropout masks are sampled at each time step for the inputs and outputs alone (no dropout is used with the recurrent connections since the use of different masks with these connections leads to deteriorated performance).
Our method and its relation to existing techniques is depicted in figure \ref{fig:depiction}. 
When used with discrete inputs (i.e.\ words) we place a distribution over the word embeddings as well. Dropout in the word-based model corresponds then to randomly dropping word \textit{types} in the sentence, and might be interpreted as forcing the model not to rely on single words for its task.

We next survey related literature and background material, and then formalise our approximate inference for the Variational RNN, resulting in the dropout variant proposed above. Experimental results are presented thereafter.

\section{Related Research}


In the past few years a considerable body of work has been collected demonstrating the negative effects of a naive application of dropout in RNNs' recurrent connections.
\citet{pachitariu2013regularization}, working with language models, reason that noise added in the recurrent connections of an RNN leads to model instabilities. Instead, they add noise to the decoding part of the model alone.
\citet{bayer2013fast} apply a deterministic approximation of dropout (fast dropout) in RNNs. They reason that with dropout, the RNN's dynamics change dramatically, and that dropout should be applied to the ``non-dynamic'' parts of the model -- connections feeding from the hidden layer to the output layer.
\citet{pham2014dropout} assess dropout with handwriting recognition tasks. They conclude that dropout in recurrent layers disrupts the RNN's ability to model sequences, and that dropout should be applied to feed-forward connections and not to recurrent connections. 
The work by \citet*{zaremba2014recurrent} was developed in parallel to \citet{pham2014dropout}. \citet{zaremba2014recurrent} assess the performance of dropout in RNNs on a wide series of tasks. They show that applying dropout to the non-recurrent connections alone results in improved performance, and provide (as yet  unbeaten) state-of-the-art results in language modelling on the Penn Treebank. 
They reason that without dropout only small models were used in the past in order to avoid overfitting, whereas with the application of dropout larger models can be used, leading to improved results.
This work is considered a reference implementation by many (and we compare to this as a baseline below).
\citet{bluche2015apply} extend on the previous body of work and perform exploratory analysis of the performance of dropout before, inside, and after the RNN's unit. They provide mixed results, not showing significant improvement on existing techniques.
More recently, and done in parallel to this work, \citet{moon2015rnndrop} suggested a new variant of dropout in RNNs in the speech recognition community. They randomly drop elements in the LSTM's internal cell $\c_t$ and use the same mask at every time step. 
This is the closest to our proposed approach (although fundamentally different to the approach we suggest, explained in \S\ref{sec:impl-details}), and we compare to this variant below as well.

Existing approaches are \textit{based on an empirical experimentation} with different flavours of dropout, following a process of trial-and-error. These approaches have led many to believe that dropout cannot be extended to a large number of parameters within the recurrent layers, leaving them with no regularisation. 
In contrast to these conclusions, we show that it is possible to derive a variational inference based variant of dropout which successfully regularises such parameters, by grounding our approach in recent theoretical research.

\section{Background}

We review necessary background in Bayesian neural networks and approximate variational inference. Building on these ideas, in the next section we propose approximate inference in the probabilistic RNN which will lead to a new variant of dropout.

\newcommand{\h}{\mathbf{h}}
\newcommand{\U}{\mathbf{U}}

\subsection{Bayesian Neural Networks}\label{sec:BNN}

Given training inputs $\X = \{ \x_1, \hdots, \x_N \}$ and their corresponding outputs $\Y = \{\y_1, \hdots, \y_N\}$, in Bayesian (parametric) regression we would like to infer parameters $\bo$ of a function $\y = \f^\bo(\mathbf{x})$ that are \textit{likely to have generated} our outputs. 
What parameters are likely to have generated our data? Following the Bayesian approach we would put some \textit{prior} distribution over the space of parameters, $p(\bo)$. This distribution represents our prior belief as to which parameters are likely to have generated our data.
We further need to define a likelihood distribution $p(\y | \x, \bo)$.
For classification tasks we may assume a softmax likelihood, 
\begin{align*}
p \big( y=d | \x, \bo \big) = \text{Categorical}\left( \exp( f_d^\bo(\x) ) / \sum_{d'} \exp( f^\bo_{d'}(\x)) \right)
\end{align*}
or a Gaussian likelihood for regression. 
Given a dataset $\X, \Y$, we then look for the \textit{posterior} distribution over the space of parameters: $p(\bo | \X, \Y)$.
This distribution captures how likely various function parameters are given our observed data.
With it we can predict an output for a new input point $\x^*$ by integrating
\begin{align} \label{eq:post}
p(\y^* | \x^*, \X, \Y) = \int p(\y^* | \x^*, \bo) p(\bo | \X, \Y) \td \bo.
\end{align}

One way to define a distribution over a parametric set of functions is to place a prior distribution over a \textit{neural network's} weights, resulting in a \textit{Bayesian NN} \citep{mackay1992practical,neal1995bayesian}.
Given weight matrices $\W_i$ and bias vectors $\Bb_i$ for layer $i$, we often place standard matrix Gaussian prior distributions over the weight matrices, $p(\W_i) = \N(\bz, \I)$
and often assume a point estimate for the bias vectors for simplicity.

\subsection{Approximate Variational Inference in Bayesian Neural Networks}
We are interested in finding the distribution of weight matrices (parametrising our functions) that have generated our data. This is the posterior over the weights given our observables $\X, \Y$: $p( \bo | \X, \Y )$. 
This posterior is not tractable in general, and we may use variational inference to approximate it (as was done in \citep{hinton1993keeping,
barber1998ensemble,
graves2011practical,
blundell2015weight}). We need to define an approximating variational distribution $q( \bo )$, and then minimise the KL divergence between the approximating distribution and the full posterior:
\begin{align} \label{eq:KL:BNN}
\KL \big( q(\bo) || p(\bo | \X, \Y ) \big)
&\propto - \int q(\bo) \log p(\Y | \X, \bo) \td \bo + \KL(q(\bo) || p(\bo)) 
\notag \\
&=- \sum_{i=1}^N \int q(\bo) \log p(\y_i | \f^\bo(\x_i)) \td \bo + \KL(q(\bo) || p(\bo)).
\end{align}

We next extend this approximate variational inference to probabilistic RNNs, and use a $q(\bo)$ distribution that will give rise to a new variant of dropout in RNNs.

\section{Variational Inference in Recurrent Neural Networks}

In this section we will concentrate on simple RNN models for brevity of notation. Derivations for LSTM and GRU follow similarly.
Given input sequence $\x = [\x_1, ..., \x_T]$ of length $T$, a simple RNN is formed by a repeated application of a function $\f_{\h}$. This generates a hidden state $\h_t$ for time step $t$:
\begin{align*}
\h_t &= \f_{\h}(\x_t, \h_{t-1}) = \sigma(\x_t \W_\h + \h_{t-1} \U_\h + \Bb_\h)
\end{align*}
for some non-linearity $\sigma$.
The model output can be defined, for example, as $\f_\y(\h_T) = \h_T \W_\y + \Bb_\y$. We view this RNN as a probabilistic model by regarding $\bo = \{\W_\h, \U_\h, \Bb_\h, \W_\y, \Bb_\y\}$ as random variables (following normal prior distributions).
To make the dependence on $\bo$ clear, we write $\f^\bo_\y$ for $\f_\y$ and similarly for $\f^\bo_\h$.
We define our probabilistic model's likelihood as above (section \ref{sec:BNN}).
The posterior over random variables $\bo$ is rather complex, and we use variational inference with approximating distribution $q(\bo)$ to approximate it.

Evaluating each sum term in eq.\ \eqref{eq:KL:BNN} above with our RNN model we get
\begin{align*}
\int q(\bo) \log p(\y | \f^\bo_\y(\h_T) ) \td \bo
&= \int q(\bo) \log p\bigg(\y \bigg| 
\f^\bo_\y \big(
  \f^\bo_\h(\x_T, \h_{T-1})
\big)
\bigg) \td \bo \\
&= \int q(\bo) \log p\bigg(\y \bigg| 
\f^\bo_\y \big(
  \f^\bo_\h(\x_T, \f^\bo_\h(... \f^\bo_\h(\x_1, \h_0) ...))
\big)
\bigg) \td \bo
\end{align*}
with $\h_0 = \textbf{0}$. We approximate this with Monte Carlo (MC) integration with a single sample:
\newcommand{\boh}{{\widehat{\bo}}}
\begin{align*}
\approx 
\log p\bigg(\y \bigg| 
\f^\boh_\y \big(
  \f^\boh_\h(\x_T, \f^\boh_\h(... \f^\boh_\h(\x_1, \h_0) ...))
\big)
\bigg), && \widehat{\bo} \sim q(\bo)
\end{align*}
resulting in an unbiased estimator to each sum term.

This estimator is plugged into equation \eqref{eq:KL:BNN} to obtain our minimisation objective
\begin{align}
\cL 
&\approx
-\sum_{i=1}^N 
\log p\bigg(\y_i \bigg| 
\f^{\boh_i}_\y \big(
  \f^{\boh_i}_\h(\x_{i,T}, \f^{\boh_i}_\h(... \f^{\boh_i}_\h(\x_{i,1}, \h_0) ...))
\big)
\bigg) + \KL(q(\bo) || p(\bo)). \label{eq:ELBO}
\end{align}
Note that for each sequence $\x_i$ we sample a new realisation $\boh_i = \{\widehat{\W}_\h^i, \widehat{\U}_\h^i, \widehat{\Bb}_\h^i, \widehat{\W}_\y^i, \widehat{\Bb}_\y^i\}$, and that each symbol in the sequence $\x_i = [\x_{i,1}, ..., \x_{i,T}]$ is passed through the function $\f^{\boh_i}_\h$ with \textit{the same weight realisations $\widehat{\W}_\h^i, \widehat{\U}_\h^i, \widehat{\Bb}_\h^i$ used at \textbf{every time step $t \leq T$}}.

Following \citep{Gal2015DropoutB} we define our approximating distribution to factorise over the weight matrices and their rows in $\bo$. For every weight matrix row $\w_k$ the approximating distribution is:
\begin{align*}
q(\w_k) &= p \N(\w_k; \textbf{0}, \sigma^2 I) + (1-p) \N(\w_k; \m_k, \sigma^2 I)
\end{align*}
with $\m_k$ variational parameter (row vector), $p$ given in advance (the dropout probability), and small $\sigma^2$. We optimise over $\m_k$ the variational parameters of the random weight matrices; these correspond to the RNN's weight matrices in the standard view\footnote{\citet{graves2013speech} further factorise the approximating distribution over the elements of each row, and use a Gaussian approximating distribution with each element (rather than a mixture); the approximating distribution above seems to give better performance, and has a close relation with dropout \citep{Gal2015DropoutB}.}. The KL in eq.\ \eqref{eq:ELBO} can be approximated as 
$L_2$ regularisation over the variational parameters $\m_k$
\citep{Gal2015DropoutB}.

Evaluating the model output $\f^{\boh}_\y(\cdot)$ with sample $\boh \sim q(\bo)$ corresponds to randomly zeroing (masking) rows in each weight matrix $\W$ during the forward pass -- i.e.\ performing dropout. 
Our objective $\cL$ is identical to that of the standard RNN.
In our RNN setting with a sequence input, each weight matrix row is randomly masked once, and importantly the same mask is used through all time steps.\footnote{In appendix \ref{sec:interp} we discuss the relation of our dropout interpretation to the ensembling one.}

Predictions can be approximated by either propagating the mean of each layer to the next (referred to as the \textit{standard dropout approximation}), or by approximating the posterior in eq.\ \eqref{eq:post} with $q(
\bo)$,
\begin{align} \label{eq:approx_predictive}
p(\y^* | \x^*, \X, \Y) &\approx
\int p ( \y^* | \x^*, \bo ) q ( \bo ) \td \bo
\approx \frac{1}{K} \sum_{k=1}^K p ( \y^* | \x^*, \widehat{\bo}_k )
\end{align}
with $\widehat{\bo}_k \sim q ( \bo )$, i.e.\ by performing dropout at test time and averaging results (\textit{MC dropout}).

\subsection{Implementation and Relation to Dropout in RNNs}
\label{sec:impl-details}

Implementing our approximate inference is identical to implementing dropout in RNNs with the \textit{same network units dropped at each time step}, randomly dropping inputs, outputs, and recurrent connections. This is in contrast to existing techniques, where different network units would be dropped at different time steps, and no dropout would be applied to the recurrent connections (fig.\ \ref{fig:depiction}). 

Certain RNN models such as LSTMs and GRUs use different \textit{gates} within the RNN units. 
For example, an LSTM is defined using four gates: ``input'', ``forget'', ``output'', and ``input modulation'',
\renewcommand{\i}{\mathbf{i}}
\newcommand{\g}{\mathbf{g}}
\renewcommand{\o}{\mathbf{o}}
\begin{align}\label{eq:non-tied-weights}
\underline{\i} &= \text{sigm} \big(\h_{t-1} \U_i + \x_t \W_i 
\big) &&
\underline{\f} = \text{sigm} \big(\h_{t-1} \U_f + \x_t \W_f 
\big) \notag \\
\underline{\o} &= \text{sigm} \big(\h_{t-1} \U_o + \x_t \W_o 
\big) && 
\underline{\g} = \text{tanh} \big(\h_{t-1} \U_g + \x_t \W_g 
\big) \notag \\
\c_t &= 
\underline{\f} \circ \c_{t-1} + \underline{\i} \circ \underline{\g} &&
\h_t =  \underline{\o} \circ \text{tanh}(\c_t)
\end{align}
with $\bo = \{ \W_i, \U_i, \W_f, \U_f, \W_o, \U_o, \W_g, \U_g\}$ weight matrices and $\circ$ the element-wise product. 
Here an internal state $\c_t$ (also referred to as \textit{cell}) is updated additively. 

Alternatively, the model could be re-parametrised as in \citep{graves2013speech}:
\begin{align}\label{eq:tied-weights}
\begin{pmatrix}
\underline{\i} \\
\underline{\f} \\
\underline{\o} \\
\underline{\g}
\end{pmatrix}
=
\begin{pmatrix}
\text{sigm} \\
\text{sigm} \\
\text{sigm} \\
\text{tanh}
\end{pmatrix}
\bigg( 
\begin{pmatrix}
\x_t \\
\h_{t-1}
\end{pmatrix}
\cdot
\W
\bigg)
\end{align}
with $\bo=\{ \W \}$, $\W$ a matrix of dimensions $2K$ by $4K$ ($K$ being the dimensionality of $\x_t$). 
We name this parametrisation a \textit{tied-weights} LSTM (compared to the \textit{untied-weights} LSTM in eq.\ \eqref{eq:non-tied-weights}).

Even though these two parametrisations result in the same \textit{deterministic} model, they lead to different approximating distributions $q(\bo)$.
With the first parametrisation one could use different dropout masks for different gates (even when the same input $\x_t$ is used). This is because the approximating distribution is placed over the matrices rather than the inputs: we might drop certain rows in one weight matrix $\W$ applied to $\x_t$ and different rows in another matrix $\W'$ applied to $\x_t$. 
With the second parametrisations we would place a distribution over the single matrix $\W$.
This leads to a faster forward-pass, but with slightly diminished results as we will see in the experiments section.

\newcommand{\bd}{\mathtt{z}}
In more concrete terms, we may write our dropout variant with the second parametrisation (eq.\ \eqref{eq:tied-weights}) as 
\begin{align}\label{eq:tied-weights-dropout}
\begin{pmatrix}
\underline{\i} \\
\underline{\f} \\
\underline{\o} \\
\underline{\g}
\end{pmatrix}
=
\begin{pmatrix}
\text{sigm} \\
\text{sigm} \\
\text{sigm} \\
\text{tanh}
\end{pmatrix}
\bigg( 
\begin{pmatrix}
\x_t \circ \bd_\x \\
\h_{t-1} \circ \bd_\h
\end{pmatrix}
\cdot
\W
\bigg)
\end{align}
with $\bd_\x, \bd_\h$ random masks repeated at all time steps (and similarly for the parametrisation in eq.\ \eqref{eq:non-tied-weights}). 



In comparison, \citet{zaremba2014recurrent}'s dropout variant replaces $\bd_\x$ in eq.\ \eqref{eq:tied-weights-dropout} with the time-dependent $\bd_\x^t$ which is sampled anew every time step (whereas $\bd_\h$ is removed and the recurrent connection $\h_{t-1}$ is not dropped):
\begin{align}\label{eq:tied-weights-naive-dropout}
\begin{pmatrix}
\underline{\i} \\
\underline{\f} \\
\underline{\o} \\
\underline{\g}
\end{pmatrix}
=
\begin{pmatrix}
\text{sigm} \\
\text{sigm} \\
\text{sigm} \\
\text{tanh}
\end{pmatrix}
\bigg( 
\begin{pmatrix}
\x_t \circ \bd_\x^t \\
\h_{t-1}
\end{pmatrix}
\cdot
\W
\bigg).
\end{align}
On the other hand, \citet{moon2015rnndrop}'s dropout variant changes eq.\ \eqref{eq:non-tied-weights} by adapting the internal cell 
\begin{align}\label{eq:tied-weights-rnn-dropout}
\c_t = \c_t \circ \bd_\c
\end{align}
with the same mask $\bd_\c$ used at all time steps. Note that unlike \citep{moon2015rnndrop}, by viewing dropout as an operation over the weights our technique trivially extends to RNNs and GRUs.



\subsection{Word Embeddings Dropout}
In datasets with continuous inputs we often apply dropout to the input layer -- i.e.\ to the input vector itself. This is equivalent to placing a distribution over the weight matrix which follows the input and approximately integrating over it (the matrix is optimised, therefore prone to overfitting otherwise). 

But for models with discrete inputs such as words (where every word is mapped to a continuous vector -- a \textit{word embedding}) this is seldom done.
With word embeddings the input can be seen as either the word embedding itself, or, more conveniently, as a ``one-hot'' encoding (a vector of zeros with $1$ at a single position). 
The product of the one-hot encoded vector with an embedding matrix $\W_E \in \mathbb{R}^{V \times D}$ (where $D$ is the embedding dimensionality and $V$ is the number of words in the vocabulary) then gives a word embedding.
Curiously, this parameter layer is the largest layer in most language applications, yet it is often not regularised. 
Since the embedding matrix is optimised it can lead to overfitting, and it is therefore desirable to apply dropout to the one-hot encoded vectors.
This in effect is identical to \textit{dropping words at random} throughout the input sentence, and can also be interpreted as encouraging the model to not ``depend'' on single words for its output.

Note that as before, we randomly set rows of the matrix $\W_E \in \mathbb{R}^{V \times D}$ to zero. Since we repeat the same mask at each time step, we drop the same words throughout the sequence -- i.e.\ we drop word types at random rather than word tokens (as an example, the sentence ``the dog and the cat'' might become ``--- dog and --- cat'' or ``the --- and the cat'', but never ``--- dog and the cat''). A possible inefficiency implementing this is the requirement to sample $V$ Bernoulli random variables, where $V$ might be large. This can be solved by the observation that for sequences of length $T$, at most $T$ embeddings could be dropped (other dropped embeddings have no effect on the model output). For $T \ll V$ it is therefore more efficient to first map the words to the word embeddings, and only then to zero-out word embeddings based on their word type.

\section{Experimental Evaluation}


We start by implementing our proposed dropout variant into the Torch implementation of \citet{zaremba2014recurrent}, that has become a reference implementation for many in the field. 
\citet{zaremba2014recurrent} have set a benchmark on the Penn Treebank that to the best of our knowledge hasn't been beaten for the past 2 years.
We improve on \citep{zaremba2014recurrent}'s results,
and show that our dropout variant improves model performance compared to early-stopping and compared to using under-specified models.
We continue to evaluate our proposed dropout variant with both LSTM and GRU models on a sentiment analysis task where labelled data is scarce. We finish by giving an in-depth analysis of the properties of the proposed method, with code and many experiments deferred to the appendix due to space constraints.

\subsection{Language Modelling}

We replicate the language modelling experiment of \citet*{zaremba2014recurrent}. 
The experiment uses the Penn Treebank, a standard benchmark in the field.
This dataset is considered a small one in the language processing community, with $887,521$ tokens (words) in total, making overfitting a considerable concern. 
Throughout the experiments we refer to LSTMs with the dropout technique proposed following our Bayesian interpretation as \textit{Variational LSTMs}, and refer to existing dropout techniques as \textit{naive dropout LSTMs} 
(eq.\ \eqref{eq:tied-weights-naive-dropout}, 
different masks at different steps, applied to the input and output of the LSTM alone). We refer to LSTMs with no dropout as \textit{standard LSTMs}. 

We implemented a Variational LSTM for both the medium model of \citep{zaremba2014recurrent}  (2 layers with 650 units in each layer) as well as their large model (2 layers with 1500 units in each layer). 
The only changes we've made to \citep{zaremba2014recurrent}'s setting are
1) using our proposed dropout variant instead of naive dropout, and
2) tuning weight decay (which was chosen to be zero in \citep{zaremba2014recurrent}).
All other hyper-parameters are kept identical to \citep{zaremba2014recurrent}: learning rate decay was not tuned for our setting and is used following \citep{zaremba2014recurrent}. 
Dropout parameters were optimised with grid search
(tying the dropout probability over the embeddings together with the one over the recurrent layers, and tying the dropout probability for the inputs and outputs together as well). These are chosen to minimise validation perplexity\footnote{Optimal probabilities are 0.3 and 0.5 respectively for the large model, compared \citep{zaremba2014recurrent}'s 0.6 dropout probability, and 0.2 and 0.35 respectively for the medium model, compared \citep{zaremba2014recurrent}'s 0.5 dropout probability.}.
We further compared to \citet{moon2015rnndrop} who only drop elements in the LSTM internal state using the same mask at all time steps (in addition to performing dropout on the inputs and outputs, eq.\ \eqref{eq:tied-weights-rnn-dropout}). 
We implemented their dropout variant with each model size, and repeated the procedure above to find optimal dropout probabilities 
(0.3 with the medium model, and 0.5  with the large model). We had to use early stopping for the large model with \citep{moon2015rnndrop}'s variant as the model starts overfitting after 16 epochs. 
\citet{moon2015rnndrop} proposed their dropout variant within the speech recognition community, where they did not have to consider embeddings overfitting (which, as we will see below, affect the recurrent layers considerably). We therefore performed an additional experiment using \citep{moon2015rnndrop}'s variant together with our embedding dropout (referred to as \textit{\citet{moon2015rnndrop}+emb dropout}).

Our results are given in table \ref{table:LM}. 
For the variational LSTM we give results using both the tied weights model (eq.\ \eqref{eq:tied-weights}--\eqref{eq:tied-weights-dropout}, \textit{Variational (tied weights)}), and without weight tying (eq.\ \eqref{eq:non-tied-weights}, \textit{Variational (untied weights)}). For each model we report performance using both the standard dropout approximation (averaging the weights at test time -- propagating the mean of each approximating distribution as input to the next layer), and using MC dropout (obtained by performing dropout at test time 1000 times, and averaging the model outputs following eq.\ \eqref{eq:approx_predictive}, denoted \textit{MC}).
For each model we report average perplexity and standard deviation (each experiment was repeated 3 times with different random seeds and the results were averaged).
Model training time is given in \textit{words per second} (WPS).

It is interesting that using the dropout approximation, weight tying results in lower validation error and test error than the untied weights model. But with MC dropout the untied weights model performs much better.
Validation perplexity for the large model is improved from \citep{zaremba2014recurrent}'s $82.2$ down to $77.3$ (with weight tying), or $77.9$ without weight tying. Test perplexity is reduced from $78.4$ down to $73.4$ (with MC dropout and untied weights). 
To the best of our knowledge, these are currently the best single model perplexities on the Penn Treebank.

It seems that \citet{moon2015rnndrop} underperform even compared to \citep{zaremba2014recurrent}. With no embedding dropout the large model overfits and early stopping is required (with no early stopping the model's validation perplexity goes up to 131 within 30 epochs). Adding our embedding dropout, the model performs much better, but still underperforms compared to applying dropout on the inputs and outputs alone.

Comparing our results to the non-regularised LSTM (evaluated with early stopping, giving similar performance as the early stopping experiment in \citep{zaremba2014recurrent}) we see that for either model size an improvement can be obtained by using our dropout variant. 
Comparing the medium sized Variational model to the large one we see that a significant reduction in perplexity can be achieved by using a larger model. This cannot be done with the non-regularised LSTM, where a larger model leads to worse results. This shows that reducing the complexity of the model, a possible approach to avoid overfitting, actually leads to a worse fit when using dropout. 

\begin{table}[t!]
\center
\def\arraystretch{1.2}
\hspace*{-3mm}
\tabcolsep=0.14cm
\begin{tabular}{c|ccc|ccc}
 & \multicolumn{3}{c}{\small Medium LSTM} & \multicolumn{3}{c}{\small Large LSTM} \\ 
 & Validation & Test & WPS & Validation & Test & WPS \\ 
\hline 
Non-regularized (early stopping) & 
$121.1$ & $121.7$ & $5.5$K & $128.3$ & $127.4$ & $2.5$K \\  
\citet{moon2015rnndrop} & $100.7$ & $97.0$ & $4.8$K & $122.9$ & $118.7$ & $3$K \\  
\citet{moon2015rnndrop} +emb dropout & $88.9$ & $86.5$ & $4.8$K & $88.8$ & $86.0$ & $3$K \\  
\citet{zaremba2014recurrent} & $86.2$ & $82.7$ & $5.5$K & $82.2$ & $78.4$ & $2.5$K \\  
\hline 
Variational (tied weights) & $81.8 \pm 0.2$ & $79.7 \pm 0.1$ & $4.7$K & $77.3 \pm 0.2$ & $75.0 \pm 0.1$ & $2.4$K \\  
Variational (tied weights, MC) & $-$ & $79.0 \pm 0.1$ & $-$ & $-$ & $74.1 \pm 0.0$ & $-$ \\  
Variational (untied weights) & $81.9 \pm 0.2$ & $79.7 \pm 0.1$ & $2.7$K & $77.9 \pm 0.3$ & $75.2 \pm 0.2$ & $1.6$K \\ 
Variational (untied weights, MC) & $-$ & $\mathbf{78.6 \pm 0.1}$ & $-$ & $-$ & $\mathbf{73.4 \pm 0.0}$ & $-$ \\  
\hline
\end{tabular}
\vspace{2mm}
\caption{Single model perplexity (on test and validation sets) for the Penn Treebank language modelling task. Two model sizes are compared (a medium and a large LSTM, following \citep{zaremba2014recurrent}'s setup), with number of processed words per second (WPS) reported. 
Both dropout approximation and MC dropout are given for the test set with the Variational model.
A common approach 
for regularisation 
is to reduce model complexity (necessary with the non-regularised LSTM).
With the Variational models however, a significant reduction in perplexity is achieved by using larger models. 
}
\label{table:LM}
\vspace{-9mm}
\end{table}



We also see that the tied weights model achieves very close performance to that of the untied weights one when using the dropout approximation. Assessing model run time though (on a Titan X GPU), we see that tying the weights results in a more time-efficient implementation. This is because the single matrix product is implemented as a single GPU kernel, instead of the four smaller matrix products used in the untied weights model (where four GPU kernels are called sequentially). Note though that a low level implementation should give similar run times.

%
We further experimented with a model averaging experiment following \citep{zaremba2014recurrent}'s setting, where several large models are trained independently with their outputs averaged.
We used Variational LSTMs with MC dropout following the setup above.
Using 10 Variational LSTMs we improve \citep{zaremba2014recurrent}'s test set perplexity from $69.5$ to $68.7$ -- obtaining identical perplexity to \citep{zaremba2014recurrent}'s experiment with 38 models. 

\begin{figure}[b!]
\vspace{-8mm}
\includegraphics[width=\linewidth]{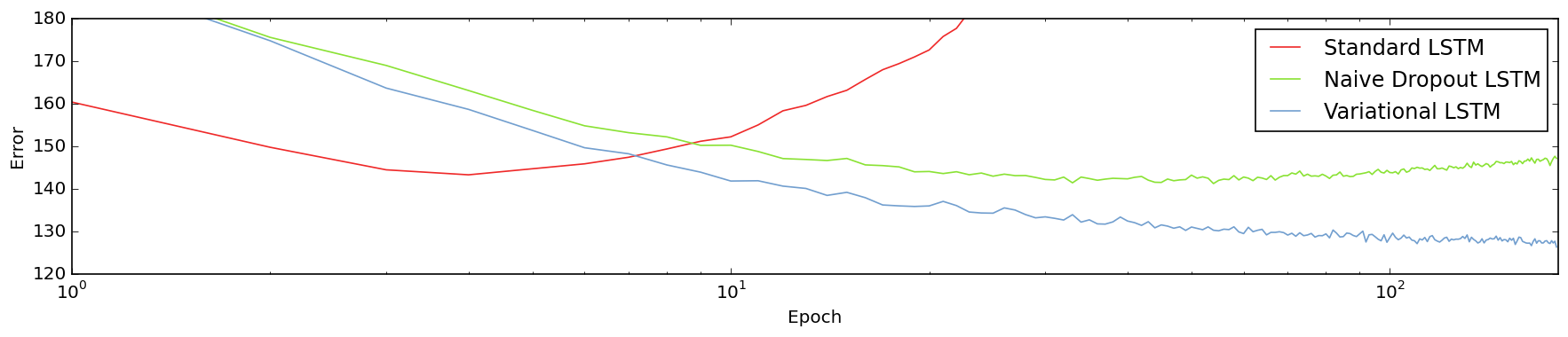}
\vspace{-7mm}
\caption{Medium model validation perplexity for the Penn Treebank language modelling task. Learning rate decay was reduced to assess model overfitting using dropout alone. Even with early stopping, Variational LSTM achieves lower perplexity than naive dropout LSTM and standard LSTM. 
Lower perplexity for all models can be achieved with learning rate decay scheduling, seen in table \ref{table:LM}.}
\label{fig:LM}
\vspace{2mm}
\end{figure}

Lastly, we report validation perplexity with reduced learning rate decay (with the medium model). 
Learning rate decay is often used for regularisation by setting the optimiser to make smaller steps when the model starts overfitting (as done in \citep{zaremba2014recurrent}).
By removing it we can assess the regularisation effects of dropout alone.
As can be seen in fig.\ \ref{fig:LM}, 
even with early stopping, Variational LSTM achieves lower perplexity than naive dropout LSTM and standard LSTM. Note though that a significantly lower perplexity for all models can be achieved with learning rate decay scheduling as seen in table \ref{table:LM}

\subsection{Sentiment Analysis}

We next evaluate our dropout variant with both LSTM and GRU models on a sentiment analysis task, where labelled data is scarce. We use MC dropout (which we compare to the dropout approximation further in appendix \ref{sec:exps}), and untied weights model parametrisations. 

We use the raw Cornell film reviews corpus collected by \citet{pang2005seeing}.
The dataset is composed of 5000 film reviews.
We extract consecutive segments of $T$ words from each review for $T=200$, and use the corresponding film score as the observed output $y$. The model is built from one embedding layer (of dimensionality $128$), one LSTM layer (with $128$ network units for each gate; GRU setting is built similarly), and finally a fully connected layer applied to the last output of the LSTM (resulting in a scalar output).
We use the Adam optimiser \citep{kingma2014adam} throughout the experiments, with batch size 128, and MC dropout at test time with 10 samples. 

\begin{figure*}[t!]
\captionsetup[subfigure]{justification=centering}
\begin{subfigure}[b]{0.32\linewidth}
\includegraphics[width=\linewidth]{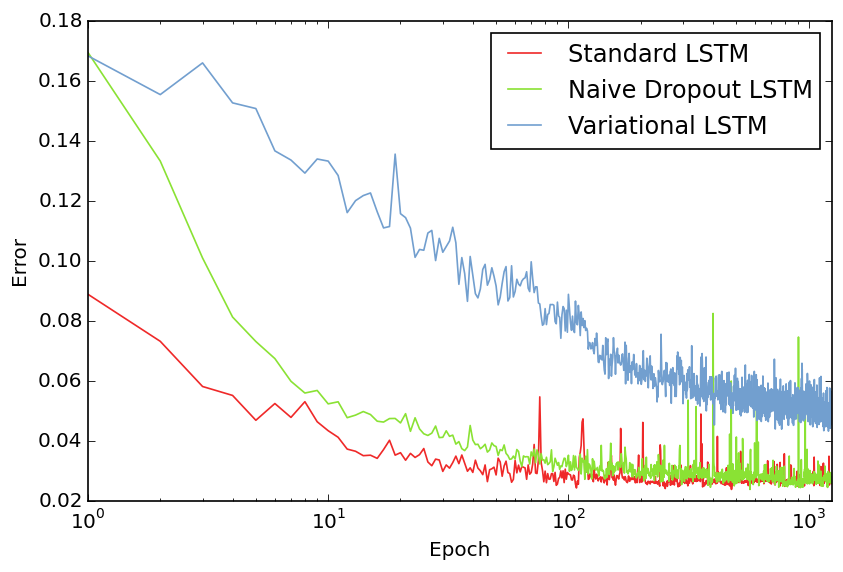}

\caption{LSTM train error: variational, naive dropout, and standard LSTM.}
\label{fig:BLSTM-a}
\end{subfigure}
\begin{subfigure}[b]{0.32\linewidth}
\includegraphics[width=\linewidth]{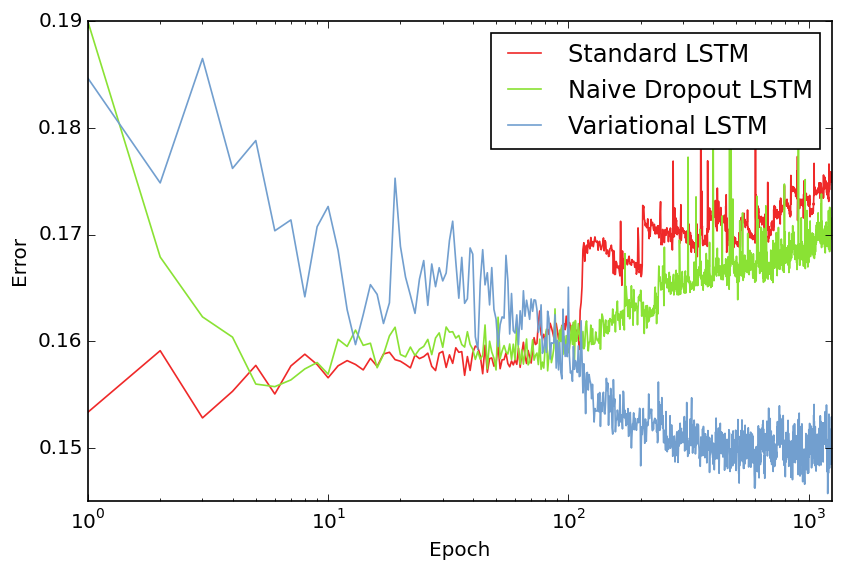}

\caption{LSTM test error: variational, naive dropout, and standard LSTM.}
\label{fig:BLSTM-b}
\end{subfigure}
\begin{subfigure}{0.32\linewidth}
\vspace{-38mm}
\includegraphics[width=\linewidth]{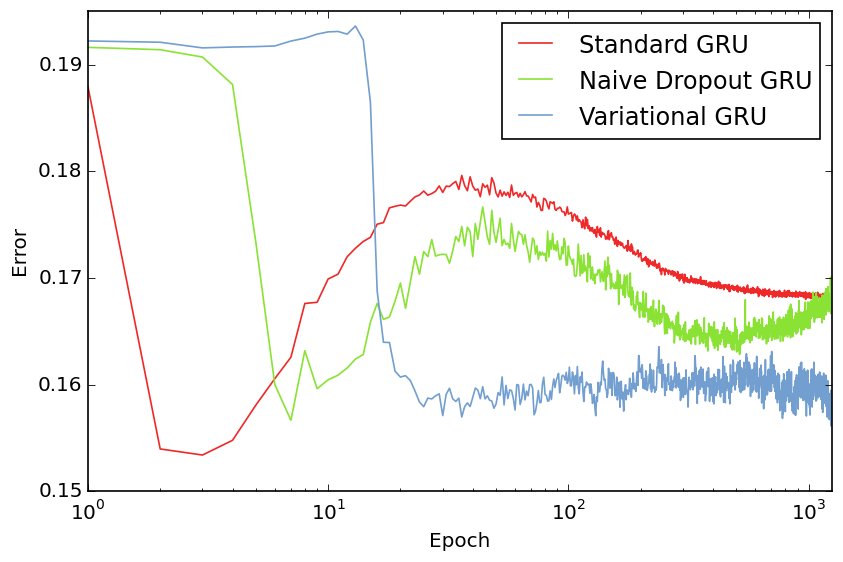}

\caption{GRU test error: variational, naive dropout, and standard LSTM.}
\label{fig:BGRU-a}
\end{subfigure}

\vspace{4mm}
\caption{
Sentiment analysis error for \textit{Variational LSTM / GRU} compared to \textit{naive dropout LSTM / GRU} and \textit{standard LSTM / GRU} (with no dropout).
}
\label{fig:BRNN}
\vspace{-4mm}
\end{figure*}

The main results can be seen in fig.\ \ref{fig:BRNN}. 
We compared Variational LSTM (with our dropout variant applied with each weight layer) to standard techniques in the field. 
Training error is shown in fig.\ \ref{fig:BLSTM-a} and test error is shown in fig.\ \ref{fig:BLSTM-b}.
Optimal dropout probabilities and weight decay were used for each model (see appendix \ref{sec:exps}). 
It seems that the only model not to overfit is the Variational LSTM, which achieves lowest test error as well. 
Variational GRU test error is shown in fig.\ \ref{fig:BGRU-a} (with loss plot given in appendix \ref{sec:exps}).
Optimal dropout probabilities and weight decay were used again for each model. 
Variational GRU avoids overfitting to the data and converges to the lowest test error. Early stopping in this dataset will result in smaller test error though (lowest test error is obtained by the non-regularised GRU model at the second epoch). 
It is interesting to note that standard techniques exhibit peculiar behaviour where test error repeatedly decreases and increases. This behaviour is not observed with the Variational GRU. 
Convergence plots of the loss for each model are given in appendix \ref{sec:exps}.

\begin{figure*}[b]
\captionsetup[subfigure]{justification=centering}
\begin{subfigure}{0.32\linewidth}
\includegraphics[width=\linewidth]{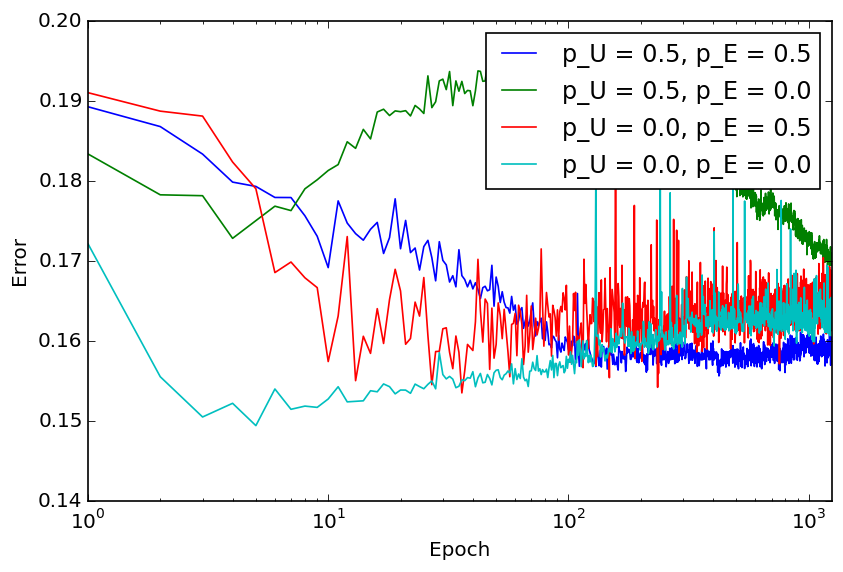}

\caption{Combinations of $p_E=0, 0.5$ with $p_U=0, 0.5$.}
\label{fig:BRNN-combinations}
\end{subfigure}
\begin{subfigure}{0.32\linewidth}
\includegraphics[width=\linewidth]{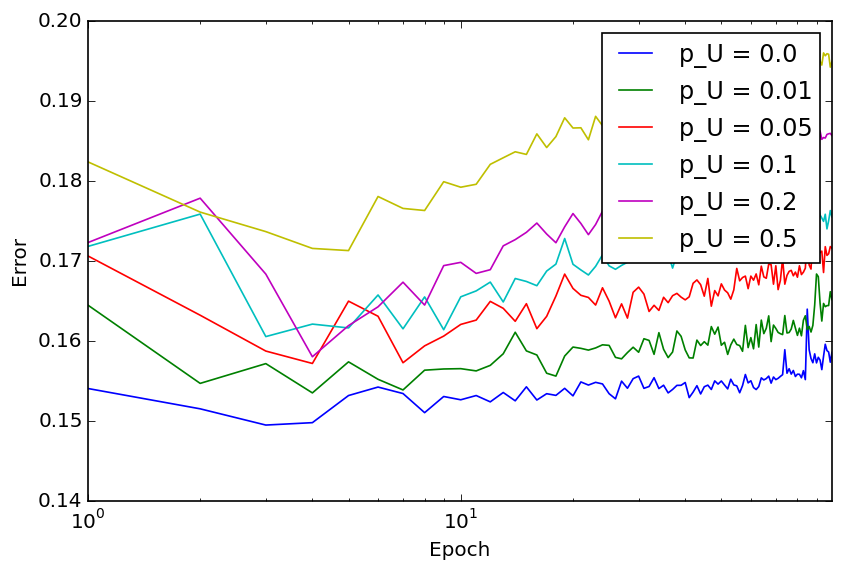}

\caption{$p_U=0, ..., 0.5$ with \\
fixed $p_E=0$.}
\label{fig:BRNN-funny1}
\end{subfigure}
\begin{subfigure}{0.32\linewidth}
\includegraphics[width=\linewidth]{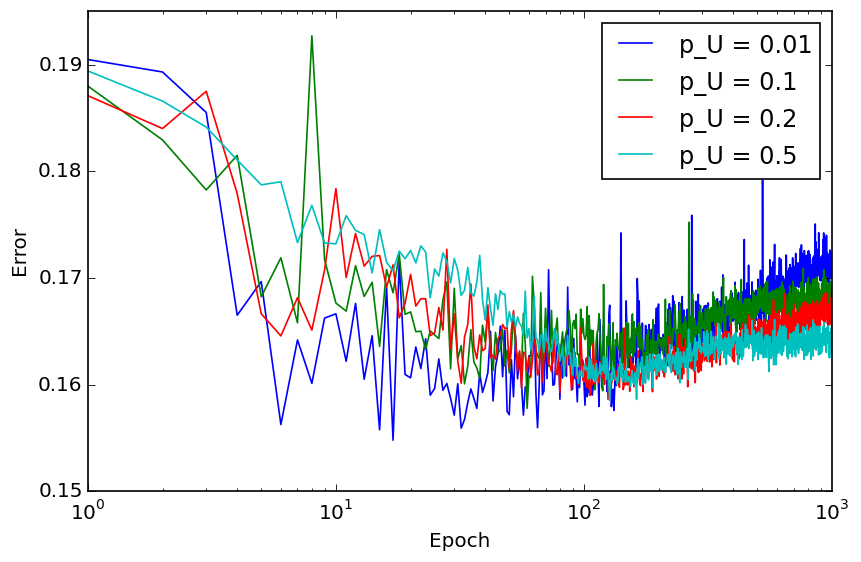}

\caption{$p_U=0, ..., 0.5$ with \\
fixed $p_E=0.5$.}
\label{fig:BRNN-funny3}
\end{subfigure}

\vspace{4mm}
\caption{Test error for Variational LSTM with various settings on the sentiment analysis task. Different dropout probabilities are used with the recurrent layer ($p_U$) and embedding layer ($p_E$).}
\vspace{2mm}
\end{figure*}

We next explore the effects of dropping-out different parts of the model. 
We assessed our Variational LSTM with different combinations of dropout over the embeddings  ($p_E=0, 0.5$) and recurrent layers ($p_U=0, 0.5$) on the sentiment analysis task. The convergence plots can be seen in figure \ref{fig:BRNN-combinations}. It seems that without both strong embeddings regularisation and strong regularisation over the recurrent layers the model would overfit rather quickly. The behaviour when $p_U=0.5$ and $p_E=0$ is quite interesting: test error decreases and then increases before decreasing again. Also, it seems that when $p_U=0$ and $p_E=0.5$ the model becomes very erratic.

Lastly, we tested the performance of Variational LSTM with different recurrent layer dropout probabilities, fixing the embedding dropout probability at either $p_E=0$ or $p_E=0.5$ (figs.\ \ref{fig:BRNN-funny1}-\ref{fig:BRNN-funny3}). These results are rather intriguing. In this experiment all models have converged, with the loss getting near zero (not shown).
Yet it seems that with no embedding dropout, a higher dropout probability within the recurrent layers leads to overfitting! This presumably happens because of the large number of parameters in the embedding layer which is not regularised. Regularising the embedding layer with dropout probability $p_E=0.5$ we see that a higher recurrent layer dropout probability indeed leads to increased \textit{robustness} to overfitting, as expected. These results suggest that embedding dropout can be of crucial importance in some tasks.

In appendix \ref{sec:exps} we assess the importance of weight decay with our dropout variant. Common practice is to remove weight decay with naive dropout. Our results suggest that weight decay plays an important role with our variant (it corresponds to our prior belief of the distribution over the weights). 

%
%

\section{Conclusions}
We presented a new technique for recurrent neural network regularisation. Our RNN dropout variant is theoretically motivated and its effectiveness was empirically demonstrated.
In future research we aim to assess model uncertainty in Variational LSTMs \citep{Gal2015DropoutB}. Together with the developments presented here, this will have important implications for modelling language ambiguity and modelling dynamics in control tasks.

%


{
\small
\setlength{\bibsep}{3pt plus 0.3ex}
\renewcommand{\baselinestretch}{0.98}
\bibliographystyle{unsrtnat}
\bibliography{example_paper}

\begin{thebibliography}{30}
\providecommand{\natexlab}[1]{#1}
\providecommand{\url}[1]{\texttt{#1}}
\expandafter\ifx\csname urlstyle\endcsname\relax
  \providecommand{\doi}[1]{doi: #1}\else
  \providecommand{\doi}{doi: \begingroup \urlstyle{rm}\Url}\fi

\bibitem[Sundermeyer et~al.(2012)Sundermeyer, Schl{\"u}ter, and
  Ney]{sundermeyer2012lstm}
Martin Sundermeyer, Ralf Schl{\"u}ter, and Hermann Ney.
\newblock {LSTM} neural networks for language modeling.
\newblock In \emph{INTERSPEECH}, 2012.

\bibitem[Kalchbrenner and Blunsom(2013)]{kalchbrenner2013recurrent}
Nal Kalchbrenner and Phil Blunsom.
\newblock Recurrent continuous translation models.
\newblock In \emph{EMNLP}, 2013.

\bibitem[Sutskever et~al.(2014)Sutskever, Vinyals, and
  Le]{sutskever2014sequence}
Ilya Sutskever, Oriol Vinyals, and Quoc~VV Le.
\newblock Sequence to sequence learning with neural networks.
\newblock In \emph{NIPS}, 2014.

\bibitem[Zaremba et~al.(2014)Zaremba, Sutskever, and
  Vinyals]{zaremba2014recurrent}
Wojciech Zaremba, Ilya Sutskever, and Oriol Vinyals.
\newblock Recurrent neural network regularization.
\newblock \emph{arXiv preprint arXiv:1409.2329}, 2014.

\bibitem[Hinton(2012)]{hinton2012improving}
Geoffrey E~others Hinton.
\newblock Improving neural networks by preventing co-adaptation of feature
  detectors.
\newblock \emph{arXiv preprint arXiv:1207.0580}, 2012.

\bibitem[Srivastava et~al.(2014)Srivastava, Hinton, Krizhevsky, Sutskever, and
  Salakhutdinov]{srivastava2014dropout}
Nitish Srivastava, Geoffrey Hinton, Alex Krizhevsky, Ilya Sutskever, and Ruslan
  Salakhutdinov.
\newblock Dropout: A simple way to prevent neural networks from overfitting.
\newblock \emph{JMLR}, 2014.

\bibitem[Pachitariu and Sahani(2013)]{pachitariu2013regularization}
Marius Pachitariu and Maneesh Sahani.
\newblock Regularization and nonlinearities for neural language models: when
  are they needed?
\newblock \emph{arXiv preprint arXiv:1301.5650}, 2013.

\bibitem[Bayer et~al.(2013)Bayer, Osendorfer, Korhammer, Chen, Urban, and
  van~der Smagt]{bayer2013fast}
Justin Bayer, Christian Osendorfer, Daniela Korhammer, Nutan Chen, Sebastian
  Urban, and Patrick van~der Smagt.
\newblock On fast dropout and its applicability to recurrent networks.
\newblock \emph{arXiv preprint arXiv:1311.0701}, 2013.

\bibitem[Pham et~al.(2014)Pham, Bluche, Kermorvant, and
  Louradour]{pham2014dropout}
Vu~Pham, Theodore Bluche, Christopher Kermorvant, and Jerome Louradour.
\newblock Dropout improves recurrent neural networks for handwriting
  recognition.
\newblock In \emph{ICFHR}. IEEE, 2014.

\bibitem[Bluche et~al.(2015)Bluche, Kermorvant, and Louradour]{bluche2015apply}
Th{\'e}odore Bluche, Christopher Kermorvant, and J{\'e}r{\^o}me Louradour.
\newblock Where to apply dropout in recurrent neural networks for handwriting
  recognition?
\newblock In \emph{ICDAR}. IEEE, 2015.

\bibitem[Rezende et~al.(2014)Rezende, Mohamed, and
  Wierstra]{rezende2014stochastic}
Danilo~Jimenez Rezende, Shakir Mohamed, and Daan Wierstra.
\newblock Stochastic backpropagation and approximate inference in deep
  generative models.
\newblock In \emph{ICML}, 2014.

\bibitem[Blundell et~al.(2015)Blundell, Cornebise, Kavukcuoglu, and
  Wierstra]{blundell2015weight}
Charles Blundell, Julien Cornebise, Koray Kavukcuoglu, and Daan Wierstra.
\newblock Weight uncertainty in neural network.
\newblock In \emph{ICML}, 2015.

\bibitem[Hernandez-Lobato and Adams(2015)]{hernandez2015probabilistic}
Jose~Miguel Hernandez-Lobato and Ryan Adams.
\newblock Probabilistic backpropagation for scalable learning of {B}ayesian
  neural networks.
\newblock In \emph{ICML}, 2015.

\bibitem[Gal and Ghahramani(2015{\natexlab{a}})]{Gal2015Bayesian}
Yarin Gal and Zoubin Ghahramani.
\newblock Bayesian convolutional neural networks with {B}ernoulli approximate
  variational inference.
\newblock \emph{arXiv:1506.02158}, 2015{\natexlab{a}}.

\bibitem[Kingma et~al.(2015)Kingma, Salimans, and Welling]{Kingma2015}
Diederik Kingma, Tim Salimans, and Max Welling.
\newblock Variational dropout and the local reparameterization trick.
\newblock In \emph{NIPS}. Curran Associates, Inc., 2015.

\bibitem[Korattikara~Balan et~al.(2015)Korattikara~Balan, Rathod, Murphy, and
  Welling]{Murphy2015}
Anoop Korattikara~Balan, Vivek Rathod, Kevin~P Murphy, and Max Welling.
\newblock Bayesian dark knowledge.
\newblock In \emph{NIPS}. Curran Associates, Inc., 2015.

\bibitem[Gal and Ghahramani(2015{\natexlab{b}})]{Gal2015DropoutB}
Yarin Gal and Zoubin Ghahramani.
\newblock Dropout as a {B}ayesian approximation: Representing model uncertainty
  in deep learning.
\newblock \emph{arXiv:1506.02142}, 2015{\natexlab{b}}.

\bibitem[Hochreiter and Schmidhuber(1997)]{hochreiter1997long}
Sepp Hochreiter and J{\"u}rgen Schmidhuber.
\newblock Long short-term memory.
\newblock \emph{Neural computation}, 9\penalty0 (8), 1997.

\bibitem[Cho et~al.(2014)]{cho2014Learning}
Kyunghyun Cho et~al.
\newblock Learning phrase representations using {RNN} encoder--decoder for
  statistical machine translation.
\newblock In \emph{EMNLP}, Doha, Qatar, October 2014. ACL.

\bibitem[Moon et~al.(December 2015)Moon, Choi, Lee, and Song]{moon2015rnndrop}
Taesup Moon, Heeyoul Choi, Hoshik Lee, and Inchul Song.
\newblock {RnnDrop: A Novel Dropout for RNNs in ASR}.
\newblock In \emph{ASRU Workshop}, December 2015.

\bibitem[MacKay(1992)]{mackay1992practical}
David~JC MacKay.
\newblock A practical {B}ayesian framework for backpropagation networks.
\newblock \emph{Neural computation}, 4\penalty0 (3):\penalty0 448--472, 1992.

\bibitem[Neal(1995)]{neal1995bayesian}
Radford~M Neal.
\newblock \emph{Bayesian learning for neural networks}.
\newblock PhD thesis, University of Toronto, 1995.

\bibitem[Hinton and Van~Camp(1993)]{hinton1993keeping}
Geoffrey~E Hinton and Drew Van~Camp.
\newblock Keeping the neural networks simple by minimizing the description
  length of the weights.
\newblock In \emph{COLT}, pages 5--13. ACM, 1993.

\bibitem[Barber and Bishop(1998)]{barber1998ensemble}
David Barber and Christopher~M Bishop.
\newblock Ensemble learning in {B}ayesian neural networks.
\newblock \emph{NATO ASI SERIES F COMPUTER AND SYSTEMS SCIENCES}, 168:\penalty0
  215--238, 1998.

\bibitem[Graves(2011)]{graves2011practical}
Alex Graves.
\newblock Practical variational inference for neural networks.
\newblock In \emph{NIPS}, 2011.

\bibitem[Graves et~al.(2013)Graves, Mohamed, and Hinton]{graves2013speech}
Alan Graves, Abdel-rahman Mohamed, and Geoffrey Hinton.
\newblock Speech recognition with deep recurrent neural networks.
\newblock In \emph{ICASSP}. IEEE, 2013.

\bibitem[Pang and Lee(2005)]{pang2005seeing}
Bo~Pang and Lillian Lee.
\newblock Seeing stars: Exploiting class relationships for sentiment
  categorization with respect to rating scales.
\newblock In \emph{ACL}. ACL, 2005.

\bibitem[Kingma and Ba(2014)]{kingma2014adam}
Diederik Kingma and Jimmy Ba.
\newblock Adam: A method for stochastic optimization.
\newblock \emph{arXiv preprint arXiv:1412.6980}, 2014.

\bibitem[Bergstra et~al.(2010)]{bergstra+al:2010-scipy}
James Bergstra et~al.
\newblock Theano: a {CPU} and {GPU} math expression compiler.
\newblock In \emph{Proceedings of the Python for Scientific Computing
  Conference ({SciPy})}, June 2010.
\newblock Oral Presentation.

\bibitem[fchollet(2015)]{keras2015}
fchollet.
\newblock Keras.
\newblock \url{https://github.com/fchollet/keras}, 2015.

\end{thebibliography}
}
\renewcommand{\baselinestretch}{1}

\newpage
\appendix
\setlength{\belowcaptionskip}{-5pt}

\section{Bayesian versus ensembling interpretation of dropout}\label{sec:interp}

Apart from our Bayesian approximation interpretation, dropout in \textit{deep} networks can also be seen as following an ensembling interpretation \citep{srivastava2014dropout}. 
This interpretation also leads to MC dropout at test time.
But the ensembling interpretation does not determine whether the ensemble should be over the network units or the weights. 
For example, in an RNN this view will \textit{not} lead to our dropout variant, unless the ensemble is \textit{defined to tie the weights of the network} ad hoc.
This is in comparison to the Bayesian approximation view where the weight tying is forced by the probabilistic interpretation of the model.

\section{Sentiment analysis -- further experiments}\label{sec:exps}

Sentiment analysis hyper-parameters were obtained by evaluating each model with dropout probabilities $0.25$ and $0.5$, 
and weight decays ranging from $10^{-6}$ to $10^{-4}$.
The optimal setting for Variational LSTM is dropout probabilities $0.25$ and weight decay $10^{-3}$, and for naive dropout LSTM the dropout probabilities are $0.5$ (no weight decay is used in reference implementations of naive dropout LSTM \citep{zaremba2014recurrent}). 

\begin{figure*}[h]
\center
\begin{minipage}{0.3\linewidth}
\includegraphics[width=\linewidth]{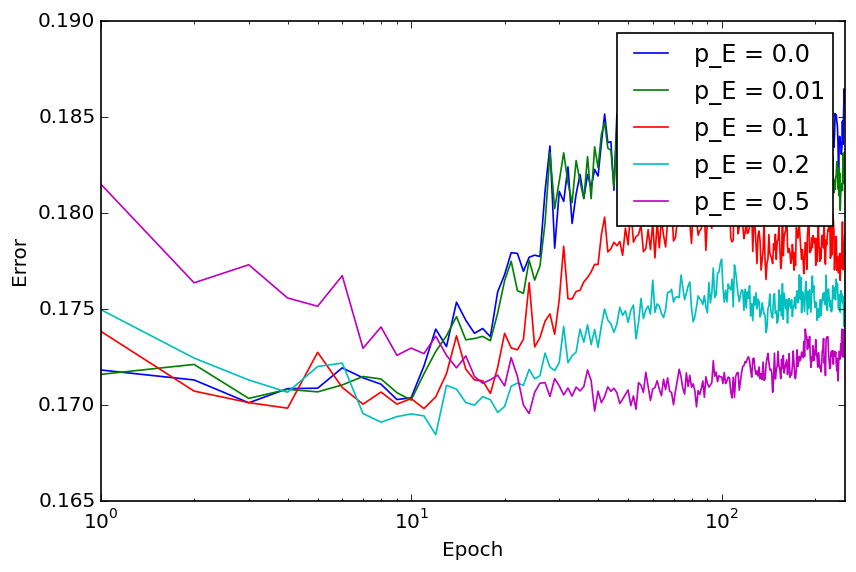}

\vspace{-2mm}
\caption{Test error for various embedding dropout probabilities, with sequence length 50.}
\label{fig:BRNN-seq-length-50}
\end{minipage}
\hspace{2mm}
\begin{minipage}{0.3\linewidth}
\includegraphics[width=\linewidth]{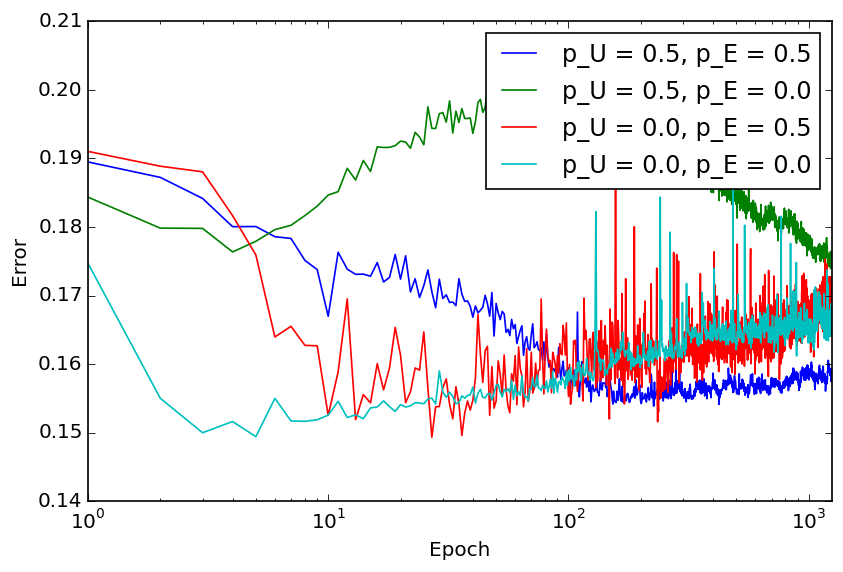}

\vspace{-2mm}
\caption{Dropout approximation in Variational LSTM with different dropout probabilities.}
\label{fig:BRNN-approx}
\end{minipage}
\end{figure*}

We assess the \textit{dropout approximation} in Variational LSTMs. The dropout approximation is often used in deep networks as means of approximating the MC estimate. In the approximation we replace each weight matrix $\M$ by $p\M$ where $p$ is the dropout probability, and perform a deterministic pass through the network (without dropping out units). This can be seen as propagating the mean of the random variables $\W$ through the network \citep{Gal2015DropoutB}. The approximation has been shown to work well for deep networks \cite{srivastava2014dropout}, yet it fails with convolution layers \citep{Gal2015Bayesian}. We assess the approximation empirically with our Variational LSTM model, repeating the first experiment with the approximation used at test time instead of MC dropout. The results can be seen in fig.\ \ref{fig:BRNN-approx}. It seems that the approximation gives a good estimate to the test error, similar to the results in figure \ref{fig:BRNN-combinations}. 

\begin{figure*}[b!]
\begin{minipage}{0.3\linewidth}
\vspace{-10mm}
\includegraphics[width=\linewidth]{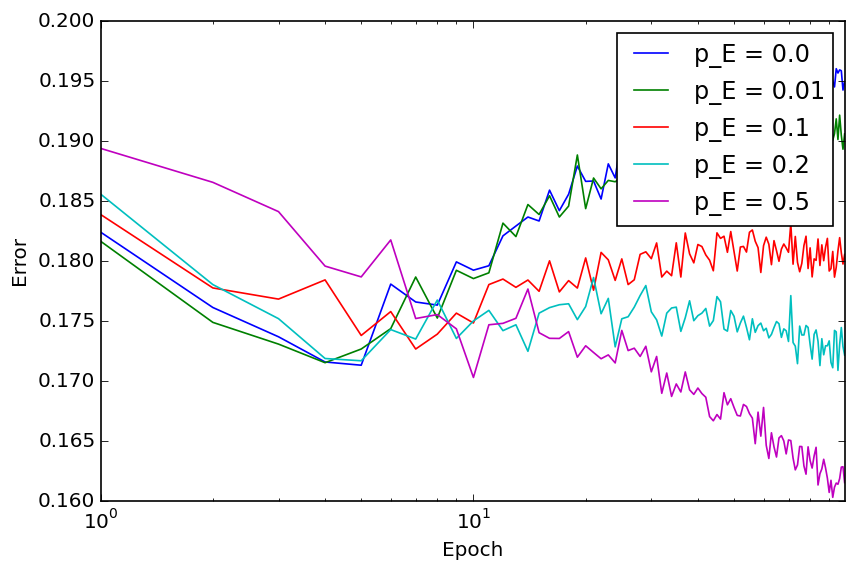}

\vspace{-2mm}
\caption{$p_E=0, ..., 0.5$ with fixed $p_U=0.5$.}
\label{fig:BRNN-seq-length-200}
\end{minipage}
\hspace{2mm}
\begin{minipage}{0.3\linewidth}
\vspace{-10mm}
\includegraphics[width=\linewidth]{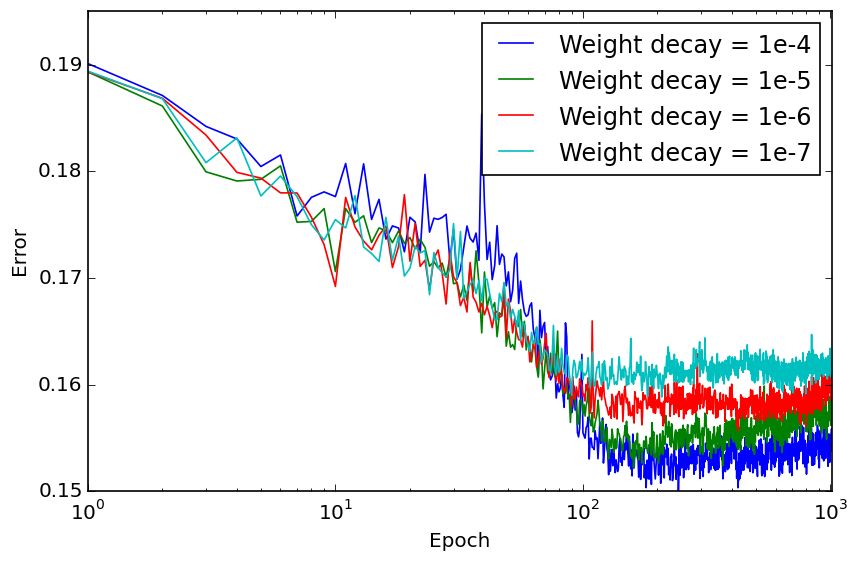}

\vspace{-2mm}
\caption{Test error for Variational LSTM with different weight decays.}
\label{fig:BRNN-weight-decay}
\end{minipage}
\hspace{2mm}
\begin{minipage}{0.3\linewidth}
\includegraphics[width=\linewidth]{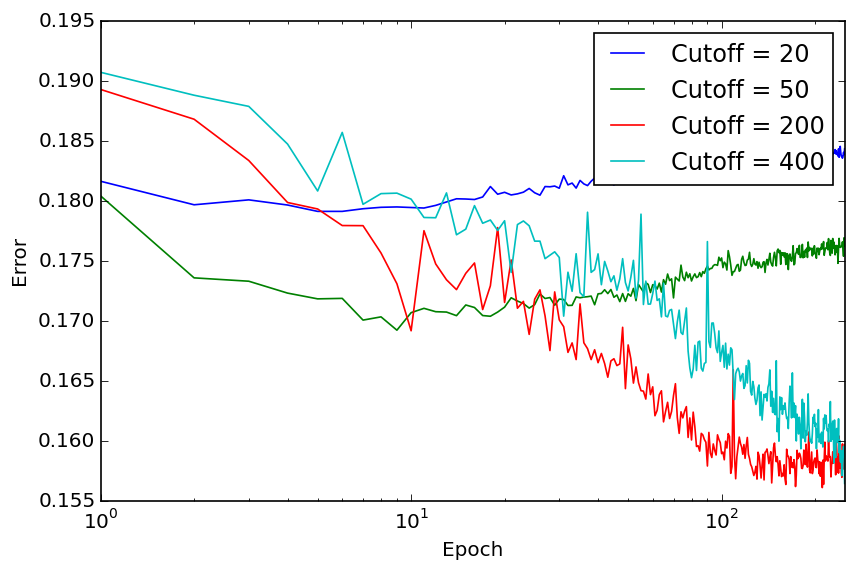}

\vspace{-2mm}
\caption{Variational LSTM test error for different sequence lengths ($T=20,50,200,400$ cut-offs).}
\label{fig:BRNN-seq-length}
\end{minipage}
\end{figure*}

We further tested the Variational LSTM model with different weight decays, observing the effects of different values for these. Note that weight decay is applied to all layers, including the embedding layer. In figure \ref{fig:BRNN-weight-decay} we can see that higher weight decay values result in lower test error, with significant differences for different weight decays. This suggests that weight decay still plays an important role even when using dropout (whereas common practice is to remove weight decay with naive dropout). Note also that the weight decay can be optimised (together with the dropout parameters) as part of the variational approximation. This is not done in practice in the deep learning community, where grid-search or Bayesian optimisation are often used for these instead.

Testing the Variational LSTM with different sequence lengths (with sequences of lengths $T=20, 50, 200, 400$) we can see that sequence length has a strong effect on model performance as well (fig.\ \ref{fig:BRNN-seq-length}). Longer sequences result in much better performance but with the price of longer convergence time. We hypothesised that the diminished performance on shorter sequences is caused by the high dropout probability on the embeddings. But a follow-up experiment with sequence lengths  50 and 200, and different embedding dropout probabilities, shows that lower dropout probabilities result in even worse model performance (figures \ref{fig:BRNN-seq-length-50} and \ref{fig:BRNN-seq-length-200}).

\begin{figure*}[h]
\vspace{-2mm}
\center
\begin{subfigure}{0.49\linewidth}
\includegraphics[width=\linewidth]{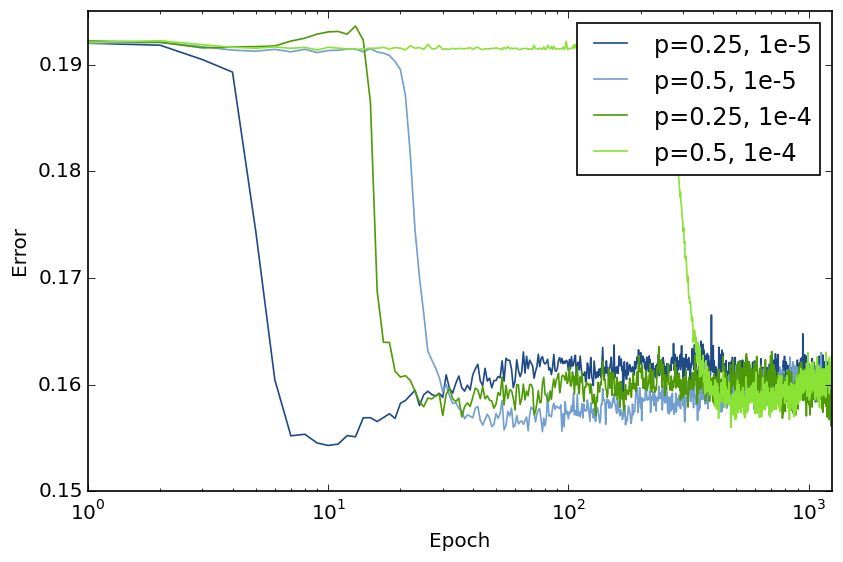}

\caption{Various Variational GRU model configurations}
\label{fig:BGRU-b}
\end{subfigure}

\vspace{3mm}
\caption{\textbf{Sentiment analysis error for \textit{Variational GRU} compared to \textit{naive dropout GRU} and \textit{standard GRU} (with no dropout).} Test error for the different models (left) and for different Variational GRU configurations (right).}
\label{fig:BGRU}
\end{figure*}

In fig.\ \ref{fig:BGRU-b} we see how different dropout probabilities and weight decays affect GRU model performance.

\begin{figure}[h!]
\center
\begin{minipage}{0.49\linewidth}
\includegraphics[width=\linewidth]{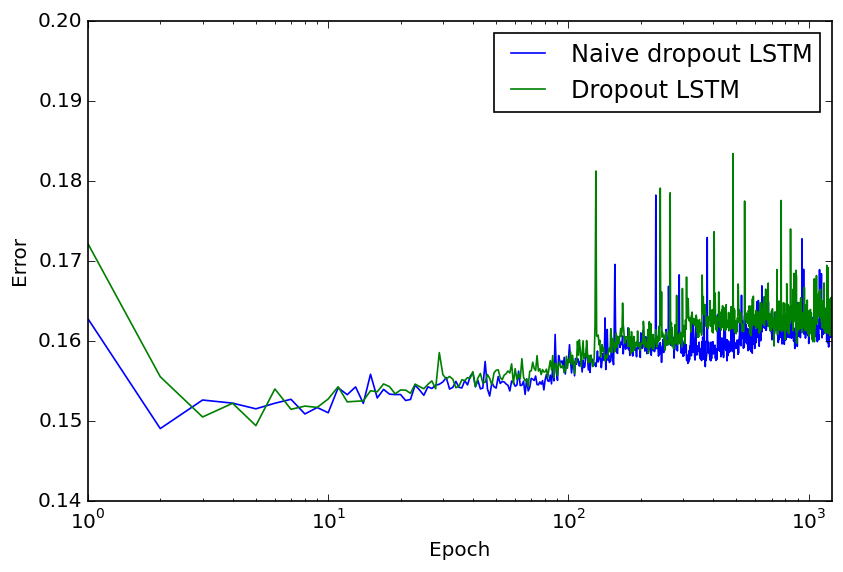}
\caption{\textit{Naive dropout LSTM} uses different dropout masks at each time step, whereas \textit{Dropout LSTM} uses the same mask at each time step. Both models apply dropout to the inputs and outputs alone, and result in identical performance.}
\label{fig:naive-dropout}
\end{minipage}
\begin{minipage}{0.49\linewidth}
\includegraphics[width=\linewidth]{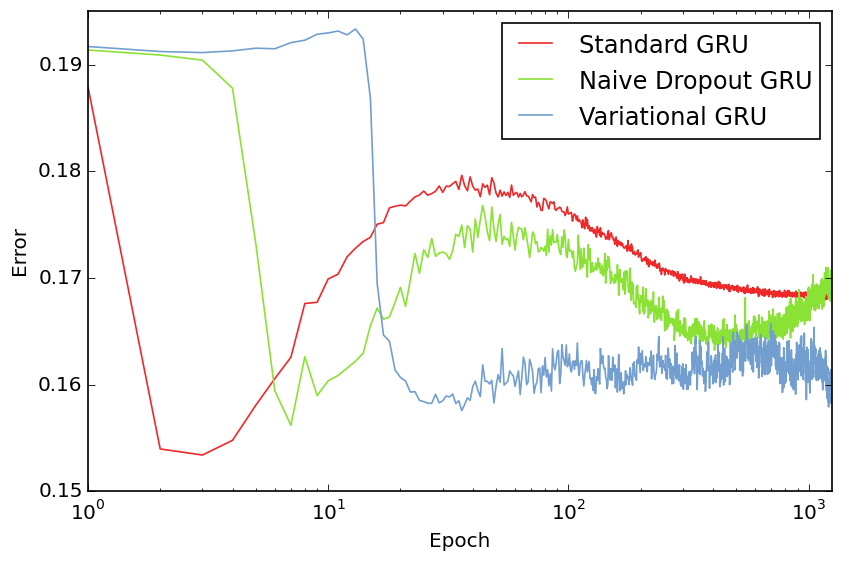}
\caption{GRU dropout approximation}
\label{fig:exp10_rmse_approx_GRU}
\end{minipage}
\end{figure}

We compare naive dropout LSTM to Variational LSTM with dropout probability in the recurrent layers set to zero: $p_U=0$ (referred to as \textit{dropout LSTM}). Both models apply dropout to the input and outputs of the LSTM alone, with no dropout applied to the embeddings. Naive dropout LSTM uses different masks at different time steps though, tied across the gates, whereas dropout LSTM uses the same mask at different time steps. 
The test error for both models can be seen in fig.\ \ref{fig:naive-dropout}.
It seems that without dropout over the recurrent layers  and embeddings both models overfit, and in fact result in identical performance.

Next, we assess the dropout approximation in the GRU model. The approximation seems to give similar results to MC dropout in the GRU model (fig.\ \ref{fig:exp10_rmse_approx_GRU}). 

Lastly, we plot the train loss for various models from the main body of the paper. All models have converged, with a stable train loss.

\begin{figure}[h!]
\center
\begin{minipage}{0.32\linewidth}
\includegraphics[width=\linewidth]{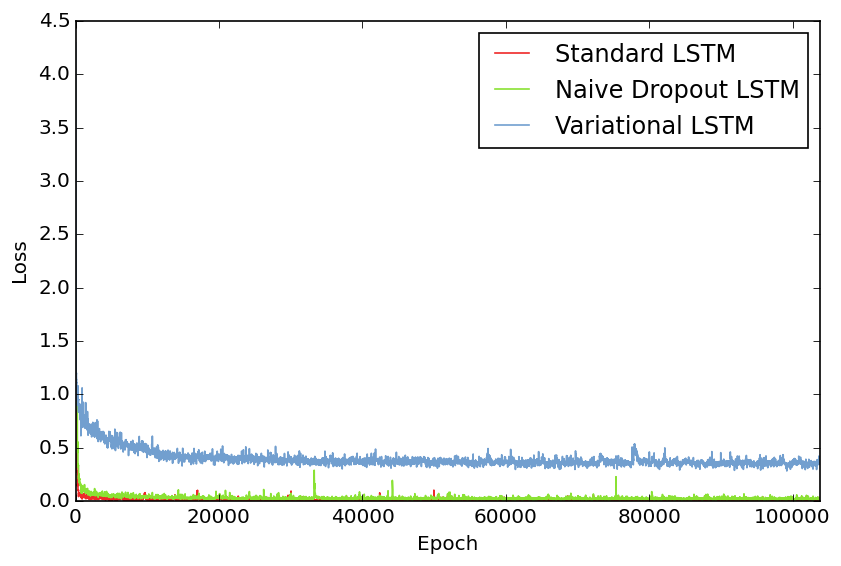}
\caption{Train loss (as a function of batches) for figure \ref{fig:BLSTM-a}}
\end{minipage}
\begin{minipage}{0.32\linewidth}
\includegraphics[width=\linewidth]{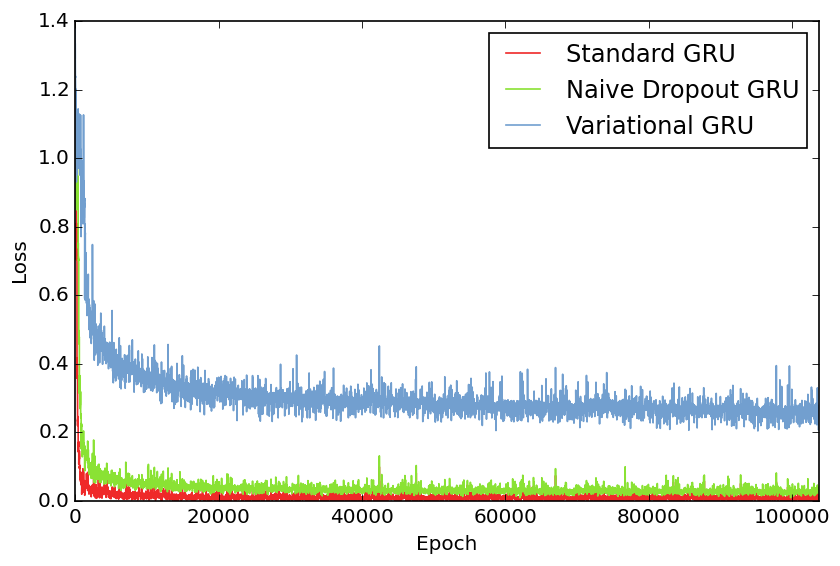}
\caption{\label{fig:BGRU-a}GRU train loss (as a function of batches) (figure \ref{fig:BGRU-a})}
\end{minipage}
\begin{minipage}{0.32\linewidth}
\includegraphics[width=\linewidth]{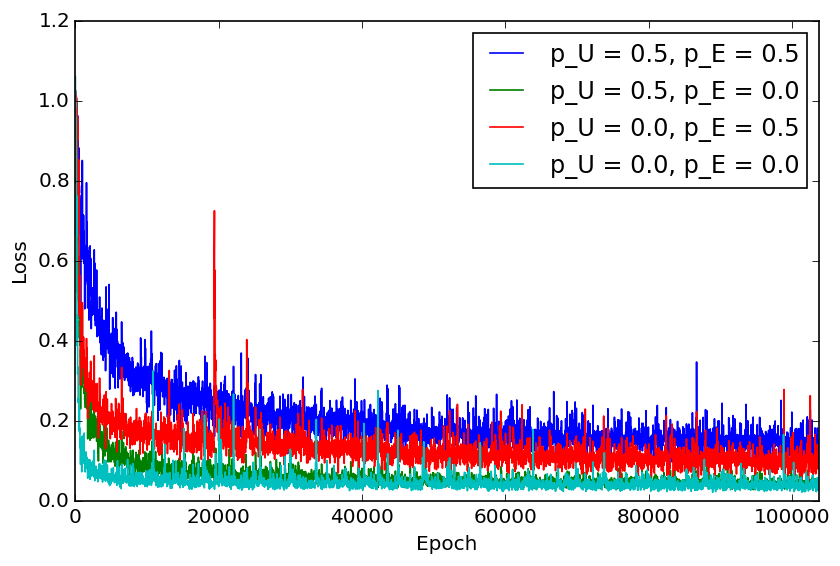}
\caption{Train loss (as a function of batches) for figure \ref{fig:BRNN-combinations}}
\end{minipage}
\end{figure}



\newpage
\section{Code}
\label{sec:code}
An efficient Theano \citep{bergstra+al:2010-scipy} implementation of the method above into Keras \citep{keras2015} is as simple as:

\begin{verbatim}
def get_output(self, train=False):
    X = self.get_input(train)

    retain_prob_W = 1. - self.p_W[0]
    retain_prob_U = 1. - self.p_U[0]
    B_W = self.srng.binomial((4, X.shape[1], self.input_dim), 
        p=retain_prob_W, dtype=theano.config.floatX)
    B_U = self.srng.binomial((4, X.shape[1], self.output_dim), 
        p=retain_prob_U, dtype=theano.config.floatX)

    xi = T.dot(X * B_W[0], self.W_i) + self.b_i
    xf = T.dot(X * B_W[1], self.W_f) + self.b_f
    xc = T.dot(X * B_W[2], self.W_c) + self.b_c
    xo = T.dot(X * B_W[3], self.W_o) + self.b_o

    [outputs, memories], updates = theano.scan(
        self._step,
        sequences=[xi, xf, xo, xc],
        outputs_info=[
            T.unbroadcast(alloc_zeros_matrix(X.shape[1], self.output_dim), 1),
            T.unbroadcast(alloc_zeros_matrix(X.shape[1], self.output_dim), 1)
        ],
        non_sequences=[self.U_i, self.U_f, self.U_o, self.U_c, B_U],
        truncate_gradient=self.truncate_gradient)

    return outputs[-1]
    
def _step(self,
          xi_t, xf_t, xo_t, xc_t,
          h_tm1, c_tm1,
          u_i, u_f, u_o, u_c, B_U):
    i_t = self.inner_activation(xi_t + T.dot(h_tm1 * B_U[0], u_i))
    f_t = self.inner_activation(xf_t + T.dot(h_tm1 * B_U[1], u_f))
    c_t = f_t * c_tm1 + i_t * self.activation(xc_t + T.dot(h_tm1 * B_U[2], u_c))
    o_t = self.inner_activation(xo_t + T.dot(h_tm1 * B_U[3], u_o))
    h_t = o_t * self.activation(c_t)
    return h_t, c_t
\end{verbatim}

\end{document}